\documentclass{article}

\usepackage{arxiv}

\usepackage[utf8]{inputenc} % allow utf-8 input
\usepackage[T1]{fontenc}    % use 8-bit T1 fonts
\usepackage{hyperref}       % hyperlinks
\usepackage{url}            % simple URL typesetting
\usepackage{booktabs}       % professional-quality tables
\usepackage{amsfonts}       % blackboard math symbols
\usepackage{nicefrac}       % compact symbols for 1/2, etc.
\usepackage{microtype}      % microtypography
\usepackage{lipsum}         % Can be removed after putting your text content
\usepackage{graphicx}
\usepackage{doi}
\usepackage{mathtools}
\usepackage[table,xcdraw]{xcolor}
\usepackage{multirow}
\usepackage{amsmath}
\usepackage{cleveref}       % smart cross-referencing
\usepackage{bbm}
\usepackage{algorithm}
\usepackage{algorithmic}
\title{DEFENDER: DTW-Based Episode Filtering Using Demonstrations for Enhancing RL Safety}

\date{}

\author{ \hspace{1mm}André Correia \\
	Universidade da Beira Interior and NOVA LINCS \\
	Covilhã, Portugal \\
	\texttt{andre.correia@ubi.pt} \\
	\And
	\hspace{1mm}Luís A. Alexandre \\
	Universidade da Beira Interior and NOVA LINCS \\
	Covilhã, Portugal \\
	\texttt{luis.alexandre@ubi.pt} \\
}

\renewcommand{\shorttitle}{DEFENDER}

\hypersetup{
pdftitle={DEFENDER: DTW-Based Episode Filtering Using Demonstrations for Enhancing RL Safety},
pdfsubject={q-bio.NC, q-bio.QM},
pdfauthor={André Correia, Luís A. Alexandre},
pdfkeywords={Demonstration, Reinforcement, Learning, Safety, Safe},
}

\begin{document}
\maketitle

\begin{abstract}
Deploying reinforcement learning agents in the real world can be challenging due to the risks associated with learning through trial and error. We propose a task-agnostic method that leverages small sets of safe and unsafe demonstrations to improve the safety of RL agents during learning. 
The method compares the current trajectory of the agent with both sets of demonstrations at every step, and filters the trajectory if it resembles the unsafe demonstrations.
We perform ablation studies on different filtering strategies and investigate the impact of the number of demonstrations on performance. Our method is compatible with any stand-alone RL algorithm and can be applied to any task. We evaluate our method on three tasks from OpenAI Gym's Mujoco benchmark and two state-of-the-art RL algorithms. The results demonstrate that our method significantly reduces the crash rate of the agent while converging to, and in most cases even improving, the performance of the stand-alone agent. 
\end{abstract}

% keywords can be removed

\section{Introduction}
\label{introduction}

Reinforcement learning (RL) \cite{barto} enables agents to learn how to behave in an environment through trial and error, but safety concerns have limited RL deployments to simulations.
Ensuring safety in unknown environments remains a challenge in RL, particularly in safety-critical domains such as healthcare and autonomous driving. 
%This is largely due to the absence of a safety mechanism that guarantees the agent avoids unsafe interactions during trial-and-error exploration.
To address these issues, safe RL studies the RL problem subject to certain constraints, with the agent aiming to maximize task reward while limiting constraint violations. 
However, deploying agents with complete knowledge of the environment to perform their tasks is unrealistic.
Alternatively, in demonstration learning (DL) the agent to learns from expert demonstrations without direct environment interaction. However, the quality of the data set plays a crucial role in the performance. Due to the difficulty of collecting a demonstration data set that covers the entire state space, pure DL policies often underperform compared to RL policies. 

In this paper, we propose a novel task-agnostic algorithm that enhances existing RL algorithms by promoting safety during interactions with the environment. Our algorithm uses a small data set of good and bad demonstrations to filter unsafe actions, terminate episodes with unsafe trajectories, and encourage the agent to explore different trajectories. 
We conduct ablation studies to evaluate filtering strategies and demonstrate the utility of our method on four tasks from OpenAI's MuJoCo environment. Our contributions in this paper are: (1) enhancing RL algorithms with safety filtering, (2) performing ablation studies on filtering strategies, (3) demonstrating the utility of our method on OpenAI's MuJoCo tasks, and (4) providing the code implementation at place to be disclosed.

\section{Related Work}
\label{sec:relatedwork}

Reinforcement learning has gained significant attention in recent years due to its wide range of applications. However, its trial-and-error nature can pose safety risks, limiting its applicability to simulation problems. 
One approach to safe reinforcement learning is to keep the agent within a safe distribution of states. For example, \cite{manifold} proposes learning a manifold that captures natural variations in the environment and uses a secondary policy to bring the agent back into the distribution of visited states.
% \cite{advantage} proposes an advantage-based mechanism for determining when the recovery policy should intervene.
Safety can be achieved through specification of constraints. For instance, \cite{barrier} proposes learning a barrier function that constrains the agent's policy to stay within a set of states that do not violate constraints. %\cite{recovery} trains two policies simultaneously to maximize expected rewards while minimizing constraint violations, where the second policy is activated when the probability of violating a constraint is above a threshold. 
Alternatively, \cite{gametheoretic} propose a zero-sum game where a second player perturbs the transition probabilities of the agent to optimize the worst transitions to produce a more robust policy. 
Some methods leverage a data set of expert demonstrations. 
\cite{ldm} proposes learning a Lyapunov function that ensures the agent's policy remains within the distribution of states of the data set. In \cite{dagger}, the authors use a pre-existing expert policy to filter the agent's action if it differs from the expert's. 
%\cite{awet} measures the DTW \cite{dtw} alignment cost of the current trajectory to the demonstrations and terminates the trajectory early if the difference is above a threshold.

\section{Preliminaries}
\label{sec:preliminaries}

\subsection{Reinforcement Learning}

Reinforcement Learning (RL) is a machine learning technique that enables an agent to learn to act in an unknown environment through trial-and-error interactions. RL is often formulated as a Markov Decision Process (MDP) described by the tuple $<S, A, R, P, \gamma>$ \cite{barto}, consisting of states $s \in S$, actions $a \in A$, transition function $P(s_{t+1}\mid s_t, a_t)$, reward function $r_t = R(s_t, a_t)$ and discount factor $\gamma$. At each timestep $t$, the agent receives the state $s_t$, selects an action $a_t = \pi(s_t)$ based on its policy, receives a reward $r_t = R(s_t, a_t)$, and transitions to a new state $s_{t+1} = P(s_t, a_t)$.
A trajectory $\tau$ is a sequence of states, actions and rewards. The goal of RL is to learn a policy $\pi$ that produces trajectories $\tau$ that maximize the expected return $\mathbb{E}_\pi[R\tau]$. Standard RL algorithms optimize a policy to maximize the expected rewards disregarding any safety concerns.

\subsection{Dynamic Time Warping}
Dynamic Time Warping (DTW) \cite{dtw} measures the similarity between two sequences of temporal data, allowing for distortions in time and variations in speed. Making it particularly useful for comparing sequences with different lengths, speeds, or underlying shapes.
The algorithm finds the optimal warping path, which minimizes the cumulative distance between corresponding points on the two sequences.
We use the alignment cost of the resulting path to measure the similarity between two sequences.

\section{Proposed Approach}
\label{sec:proposed}

\begin{algorithm}[tb]
   \caption{RL loop with DEFENDER enhancement}
   \label{alg}
\begin{algorithmic}
    \STATE {\bfseries Input:} Policy $\pi$; Memory $\beta$; Dynamics $\theta$; Constant $R_{task}$; Alignment cost functions: $Safe, Unsafe$; Task environment Env.
    \WHILE{EPISODE not DONE}
        \STATE  $a \leftarrow \pi(s)$
        \STATE  $\tau$ $\cup$ s or $(s,a)$
        \IF{$Safe(\tau) < Unsafe(\tau)$}
            \STATE $s', r, done = Env(a)$
        
        \ELSE
            \STATE $s', r, done = \theta(s, a), R_{task}, True$
        \ENDIF
        \STATE $\beta \leftarrow (s,a,r,s')$
        \STATE Optimize $\pi$ and $\theta$
    \ENDWHILE
    \STATE {\bfseries return} $\pi$
\end{algorithmic}
\end{algorithm}

%We propose a method for enhancing safety of RL agents during learning by leveraging limited demonstration data sets. RL agents estimate complex functions through trial-and-error interactions, but errors during learning can have serious consequences in the real world. 
%Demonstration learning allows agents to learn from data sets without interacting with the environment, but the policy quality is limited to the demonstrated performance and may not cover the entire state-action space, leading to compounding errors.

%To address these limitations, the proposed algorithm filters unsafe transitions using information from demonstrations to guide the RL agent and prevent errors in the environment. The algorithm can be applied to any off-the-shelf RL algorithm to enhance safety during learning without task specific constraints. 

We propose DEFENDER: DTW-Based Episode Filtering Using Demonstrations for Enhancing RL Safety, a method to improve the safety of any RL algorithm during learning by leveraging limited demonstration data sets without task-specific constraints. The algorithm keeps track of the current trajectory which is either the sequence of states or state-action pairs. We performed ablation studies to evaluate the type of trajectory which results in better performance.
We assume access to a demonstration data set $D_{demo} = \{\tau_i\}^N$ containing $N$ demonstrations, where $N > 1$ is potentially a small number that would not suffice for demonstration learning. The data set must contain both safe and unsafe trajectories. Unsafe trajectories terminated due to reaching unsafe states. Safe trajectories do not have to be perfect but must avoid unsafe states.

DEFENDER measures the alignment cost of the current trajectory at every step with each demonstration from both groups using the DTW algorithm. If the trajectory aligns better with the unsafe group the episode is terminated and the agent is discouraged from re-sampling this trajectory.
If the trajectory is terminated, the agent must be encouraged not to sample the same action that would have caused termination for the state. Otherwise, the agent will try again. A negative reward is assigned to discourage sampling the same action again. We use the lowest reward from the reward function of the respective task.

RL algorithms use the next state to perform temporal learning.
Since the filtered action is not performed, the agent does not transition to the next state. We propose predicting the next state using a dynamics model $p_\theta(s_{t+1} \mid s_t, a_t)$. The dynamics model is a simple fully-connected feed-forward network with two hidden layers parameterized by $\theta$ trained with L2 loss to imitate the true transition function of the MDP.
DEFENDER is summarized in Algorithm \ref{alg}.

\section{Filtering Strategies}

We evaluated different filtering strategies for DEFENDER. One compares the current trajectory with the complete demonstration. A second compares the trajectory with the demonstration using an equal length window. For instance, if the current trajectory has length $L$, we compare the trajectory with the last $L$ transitions of the demonstration. We also test using a fixed window of 5 and 10 transitions applied to only the trajectory, the demonstration and both. After computing the cost between the trajectory and each demonstration, we obtain a set of values for each group. We test selecting the minimum, maximum and average value from each set. This results in 24 methods to compare the trajectory with a group of demonstrations.
We can use one method to compare the trajectory with the safe demonstrations and another method to compare it with the unsafe demonstrations, resulting in 576 filtering strategies.

To determine the best filtering strategy, we trained a SAC agent on three tasks and saved the transitions of the episodes. We simulated how the agent would perform the episodes with each strategy. For each episode, we measured its length with the filter active and divide it by the length of the episode without the filter. We also determined if the filter prevents a crash. At the end, we obtained the average episode length percentages and multiplied it by the safe episode rate. We ranked the strategies by the average score for the three tasks, and selected the top 5 strategies for both state and state-action trajectories.

\section{Experiments}
\label{sec:experiments}

\begin{table}[tb]
\caption{Performance, safety and computation time of SAC and TD3 agents enhanced with our algorithm using different filters for state trajectories.}
\resizebox{\textwidth}{!}{% use resizebox with textwidth
\begin{tabular}{cc|ccc|ccc|ccc}
\hline
\multicolumn{2}{c|}{\multirow{2}{*}{Algorithm}} & \multicolumn{3}{c|}{Hopper}                        & \multicolumn{3}{c|}{InvertedDoublePendulum}       & \multicolumn{3}{c}{Walker2d}                        \\
\multicolumn{2}{c|}{}                           & Acc Reward         & \% Crash       & \% Time      & Acc Reward        & \% Crash       & \% Time      & Acc Reward          & \% Crash       & \% Time      \\ \hline
\multicolumn{2}{c|}{SAC}                        & 3474$\pm$85.0          & 64$\pm$6.0         & 100          & \textbf{9359$\pm$0.2} & 14$\pm$2.7         & 100          & 3326$\pm$254.6          & 40$\pm$4.0         & 100          \\ \hline
MinDemoW5              & MinDemoW10             & 50$\pm$4.0             & 3$\pm$0.2          & 143          & 9357$\pm$1.5          & \textbf{1$\pm$0.2} & 144          & 4149$\pm$106.3          & \textbf{3$\pm$0.6} & 157          \\
MeanBothW5             & MeanBothW5             & 3557$\pm$50.8          & 9$\pm$0.3          & \textbf{119} & 9357$\pm$0.7          & 21$\pm$0.6         & \textbf{121} & 4178$\pm$53.7           & 8$\pm$0.3          & \textbf{122} \\
MeanBothW10            & MeanBothW10            & 3500$\pm$16.4          & \textbf{8$\pm$0.7} & 138          & 9357$\pm$0.3          & 20$\pm$0.8         & 141          & 4215$\pm$76.9           & 12$\pm$0.5         & 150          \\
MeanDemoW5             & MeanDemoW10            & \textbf{3600$\pm$11.9} & 11$\pm$0.8         & 143          & 9357$\pm$0.6          & 22$\pm$1.5         & 144          & \textbf{4281$\pm$23.2}  & 13$\pm$0.1         & 156          \\
MinDemoW10             & MinDemoW10             & 1$\pm$0.0              & 0$\pm$0.0          & 152          & 9356$\pm$1.5          & \textbf{1$\pm$0.0} & 157          & 3998$\pm$105.1          & \textbf{3$\pm$0.4} & 168          \\ \hline
\multicolumn{2}{c|}{TD3}                        & 2561$\pm$1311.6        & 93$\pm$9.9         & 100          & \textbf{9355$\pm$0.2} & 47$\pm$7.9         & 100          & 4783$\pm$230.9          & 35$\pm$8.2         & 100          \\ \hline
MinDemoW5              & MinDemoW10             & 49$\pm$9.9             & \textbf{8$\pm$1.7} & 198          & 9353$\pm$0.4          & \textbf{2$\pm$0.2} & 184          & 4978$\pm$119.1          & \textbf{6$\pm$2.5} & 221          \\
MeanBothW5             & MeanBothW5             & 3652$\pm$67.4          & 16$\pm$1.9         & \textbf{143} & 9352$\pm$1.4          & 62$\pm$3.3         & \textbf{139} & 5122$\pm$30.3           & 13$\pm$4.1         & \textbf{147} \\
MeanBothW10            & MeanBothW10            & 3690$\pm$55.2          & 14$\pm$1.0         & 188          & 9352$\pm$1.1          & 55$\pm$0.5         & 177          & 5159$\pm$102.0          & 15$\pm$3.0         & 204          \\
MeanDemoW5             & MeanDemoW10            & \textbf{3821$\pm$47.1} & 15$\pm$0.2         & 198          & 9351$\pm$0.5          & 58$\pm$1.5         & 184          & \textbf{5263$\pm$138.4} & 18$\pm$2.0         & 218          \\
MinDemoW10             & MinDemoW10             & 1$\pm$0.0              & 0$\pm$0.0          & 219          & 9350$\pm$0.9          & \textbf{3$\pm$0.5} & 208          & 4571$\pm$67.5           & \textbf{5$\pm$1.6} & 242          \\ \hline
\end{tabular}
}
\label{tabstate}
\end{table}

\begin{table}[tb]
\caption{Performance, safety and computation time of SAC and TD3 agents enhanced with our algorithm using different filters for state-action trajectories.}
\resizebox{\textwidth}{!}{% use resizebox with textwidth
\begin{tabular}{cc|clc|clc|clc}
\hline
\multicolumn{2}{c|}{\multirow{2}{*}{Algorithm}} & \multicolumn{3}{c|}{Hopper}                                           & \multicolumn{3}{c|}{InvertedDoublePendulum}                     & \multicolumn{3}{c}{Walker2d}                                            \\
\multicolumn{2}{c|}{}                           & Acc Reward        & \multicolumn{1}{c}{\% Crash}       & \% Time      & Acc Reward        & \multicolumn{1}{c}{\% Crash} & \% Time      & Acc Reward         & \multicolumn{1}{c}{\% Crash}        & \% Time      \\ \hline
\multicolumn{2}{c|}{SAC}                        & 3474$\pm$85.0         & 64$\pm$6.0                             & 100          & \textbf{9359$\pm$0.2} & \textbf{14$\pm$2.7}              & 100          & 3326$\pm$254.6         & 40$\pm$4.0                              & 100          \\ \hline
MeanBothW5             & MeanBothW5             & 3475$\pm$17.7         & \multicolumn{1}{c}{9$\pm$0.8}          & \textbf{119} & 9357$\pm$0.5          & \multicolumn{1}{c}{21$\pm$1.4}   & \textbf{120} & 4207$\pm$32.8          & \multicolumn{1}{c}{\textbf{11$\pm$1.0}} & \textbf{123} \\
MeanBothW10            & MeanBothW10            & 3486$\pm$17.0         & \multicolumn{1}{c}{\textbf{8$\pm$0.1}} & 141          & 9356$\pm$0.2          & \multicolumn{1}{c}{21$\pm$1.0}   & 141          & \textbf{4261$\pm$13.0} & \multicolumn{1}{c}{14$\pm$0.4}          & 155          \\
MinTrajW5              & MinTrajW10             & 1$\pm$0.0             & \multicolumn{1}{c}{0$\pm$0.0}          & 1119         & 8$\pm$0.0             & \multicolumn{1}{c}{0$\pm$0.0}    & 728          & -4$\pm$0.0             & \multicolumn{1}{c}{0$\pm$0.0}           & 1642         \\
MinBoth                & MinTrajW10             & \textbf{3602$\pm$0.0} & \multicolumn{1}{c}{53$\pm$8.4}         & 602          & 201$\pm$4.3           & \multicolumn{1}{c}{1$\pm$0.1}    & 196          & 3878$\pm$0.0           & \multicolumn{1}{c}{24$\pm$4.3}          & 824          \\
MinBoth                & MinTrajW5              & 3443$\pm$13.1         & \multicolumn{1}{c}{11$\pm$0.2}         & 460          & 189$\pm$19.2          & \multicolumn{1}{c}{1$\pm$0.2}    & 179          & 3944$\pm$91.7          & \multicolumn{1}{c}{29$\pm$9.2}          & 621          \\ \hline
\multicolumn{2}{c|}{TD3}                        & 2561$\pm$1311.6       & 93$\pm$9.9                             & 100          & \textbf{9355$\pm$0.2} & \textbf{47$\pm$7.9}              & 100          & 4783$\pm$230.9         & 35$\pm$8.2                              & 100          \\ \hline
MeanBothW5             & MeanBothW5             & 3645$\pm$29.8         & \textbf{14$\pm$1.1}                    & \textbf{143} & 9351$\pm$0.9          & 59$\pm$12.8                      & \textbf{139} & 5196$\pm$55.0          & 17$\pm$2.4                              & \textbf{149} \\
MeanBothW10            & MeanBothW10            & 3651$\pm$35.1         & 15$\pm$1.5                             & 193          & 9351$\pm$0.3          & 58$\pm$5.2                       & 178          & \textbf{5253$\pm$78.2} & 17$\pm$1.9                              & 216          \\
MinTrajW5              & MinTrajW10             & 1$\pm$0.0             & 0$\pm$0.0                              & 2442         & 8$\pm$0.0             & 0$\pm$0.0                        & 1297         & -4$\pm$0.0             & 0$\pm$0.0                               & 3345         \\
MinBoth                & MinTrajW10             & 3759$\pm$0.0          & 60$\pm$0.3                             & 1253         & 204$\pm$7.5           & 0$\pm$0.0                        & 282          & 4991$\pm$0.0           & \textbf{13$\pm$2.2}                     & 1623         \\
MinBoth                & MinTrajW5              & \textbf{3849$\pm$0.0} & 68$\pm$0.3                             & 928          & 167$\pm$0.0           & 0$\pm$0.0                        & 251          & 5220$\pm$37.3          & \textbf{13$\pm$3.4}                     & 1196         \\ \hline
\end{tabular}
}
\label{tabstateaction}
\end{table}

\begin{table}[tb]
\caption{Performance and safety of SAC agent with DEFENDER using and MeanDemoW5 and MeanDemoW10 filters, varying number of demonstrations.}
\resizebox{\textwidth}{!}{% use resizebox with textwidth
\begin{tabular}{c|cc|cc|cc}
\hline
\multirow{2}{*}{\#Demonstrations} & \multicolumn{2}{c|}{Hopper}          & \multicolumn{2}{c|}{InvertedDoublePendulum} & \multicolumn{2}{c}{Walker2d}         \\
                                  & Acc. Reward        & \% Crash        & Acc. Reward           & \% Crash            & Acc. Reward        & \% Crash        \\ \hline
10                                & 3541$\pm$25.2          & 16$\pm$0.8          & \textbf{9356$\pm$0.2}              & \textbf{22$\pm$2.1}     & 4225$\pm$46.2          & 14$\pm$0.9          \\
20                                & 3529$\pm$13.5          & 13$\pm$1.1          & \textbf{9356$\pm$0.8}              & \textbf{22$\pm$1.7}     & 4259$\pm$26.6          & 13$\pm$1.1          \\
50                                & \textbf{3600$\pm$11.9} & \textbf{11$\pm$0.8} & \textbf{9357$\pm$0.6}     & \textbf{22$\pm$1.5}     & \textbf{4281$\pm$23.2} & \textbf{13$\pm$0.1} \\ \hline
\end{tabular}
}
\label{tabdemo}
\end{table}

In this section, we evaluate the effectiveness of our method in enhancing the safety of the underlying RL algorithms, SAC \cite{sac} and TD3 \cite{td3}. We selected three tasks from OpenAI Gym's Mujoco benchmark: Hopper, Inverted Double-Pendulum, and Walker2d. 
The episode horizon for these tasks is 1000 steps, but an episode can end early if the agent reaches an unsafe state, leading to a crash if the transition was not filtered. 
We measured the performance and safety of RL agents by accumulated reward and crash rate, respectively.
Agents were trained for 5000 episodes, and each experiment was repeated three times for seed dependency. We used a learning rate of $3^{-4}$, batch size of 256, and all the networks have 2 hidden layers with 256 neurons each. We used 50 demonstrations for both safe and unsafe groups, obtained by training a SAC agent and generating a demonstration after every episode.

We trained SAC and TD3 agents with and without our algorithm, using the top 5 filtering strategies from the ablation study for both state and state-action trajectories. Results are shown in Tables \ref{tabstate} and \ref{tabstateaction}. Our algorithm incurs extra computational cost which we present in the tables.
Overall, both SAC and TD3 agents vary consistently across filtering strategies for the same task. 
Results show that state trajectories are preferred over state-action trajectories. Not only are they less computationally expensive, but they consistently lead to far fewer crashes for the same task and filtering strategy. Some filtering strategies are too strong and prevent the agent from interacting with the environment and learning the task. Those that allow the agent to interact with the environment lead the agent to the same performance in the case of InvertedDoublePendulum and to higher performance in the case of Hopper and Walker. More importantly, the crash rate is significantly decreased with some exceptions in the InvertedDoublePendulum task. The results show that when selecting an appropriate filtering strategy, our algorithm is able to significantly reduce the crash rate and lead the agent to higher performance. However, selecting a good filtering strategy for the task may require some trial and error.

Lastly, we evaluated the impact of the number of demonstrations on the performance by training an agent with DEFENDER using 'MeanDemoW5' for the safe set, and 'MeanDemoW10' for the unsafe set as the filtering strategy, varying the number of demonstrations. Results are shown in Table \ref{tabdemo}. As expected, the larger data sets result in better crash prevention. Hence, DEFENDER is able to increase the agent's safety with small data sets, and it can be further improved by increasing the demonstration data set size.

\section{Conclusion}

In this paper, we introduced the DEFENDER algorithm that can be integrated with any RL algorithm for improving the safety of agents during learning.
Ablation studies helped identify effective filtering strategies, which we evaluated on state-of-the-art RL algorithms and multiple tasks.
The results demonstrate significant enhancement in the safety of the learning agent while maintaining or improving performance.
However, there is room for improvement, as the filtering strategy may be overly protective for certain tasks, requiring many iterations to select an appropriate strategy.
Our work highlights the potential of using demonstrations for task-agnostic enhancement of RL algorithm safety. 
We provide the code implementation of the algorithm at place to be disclosed.
%In future work, we plan to explore more robust filtering approaches to further enhance RL agent safety.

%We believe that DTW Filter can be a valuable tool for improving the safety of RL agents during trial-and-error learning, and we hope to inspire further research in this direction.

\section*{Acknowledgments}

This work is supported by NOVA LINCS (UIDB/04516/2020) with the financial support of 'FCT - Funda\c{c}\~ao para a Ci\^encia e Tecnologia' and also through the research grant '2022.14197.BD'.

\end{document}